\documentclass{article}

\usepackage{arxiv}

\usepackage[utf8]{inputenc} 
\usepackage[T1]{fontenc}    
\usepackage{hyperref}       
\usepackage{url}            
\usepackage{booktabs}       
\usepackage{amsfonts}       
\usepackage{nicefrac}       
\usepackage{microtype}      
\usepackage{cleveref}       
\usepackage{lipsum}         
\usepackage{graphicx}
\usepackage{natbib}
\usepackage{doi}

\usepackage{listings}
\usepackage{xcolor}

\definecolor{codegreen}{rgb}{0,0.6,0}
\definecolor{codegray}{rgb}{0.5,0.5,0.5}
\definecolor{codepurple}{rgb}{0.58,0,0.82}
\definecolor{backcolour}{rgb}{0.95,0.95,0.92}

\lstdefinestyle{mystyle}{
    backgroundcolor=\color{backcolour},
    commentstyle=\color{codegreen},
    keywordstyle=\color{magenta},
    numberstyle=\tiny\color{codegray},
    stringstyle=\color{codepurple},
    basicstyle=\ttfamily\footnotesize,
    breakatwhitespace=false,
    breaklines=true,
    captionpos=b,
    keepspaces=true,
    numbers=left,
    numbersep=5pt,
    showspaces=false,
    showstringspaces=false,
    showtabs=false,
    tabsize=2
}

\title{Predicting effect of novel treatments using molecular pathways and real-world data}

\newif\ifuniqueAffiliation
\uniqueAffiliationtrue

\ifuniqueAffiliation 
\author{Adrien Couetoux,  Thomas Devenyns, Lise Diagne, David Champagne, Pierre-Yves Mousset, Chris Anagnostopoulos\\
	QuantumBlack, AI by McKinsey\\
	\texttt{adrien.couetoux@mckinsey\textunderscore com}\\
}
\else
\usepackage{authblk}

\newbox{\orcid}\sbox{\orcid}{\includegraphics[scale=0.06]{orcid.pdf}}
\author[1]{%
	David Champagne, Chris Anagnostopoulos, Thomas Devenyns, Lise Diagne, Adrien Couetoux
}
\affil[1]{QuantumBlack, AI by McKinsey}
\fi

\renewcommand{\shorttitle}{Predicting effect of novel treatments using molecular pathways and real-world data}

\hypersetup{
pdftitle={Predicting effect of novel treatments using molecular pathways and real-world data},
pdfsubject={q-bio.NC, q-bio.QM},
pdfauthor={David Champagne, Chris Anagnostopoulos, Thomas Devenyns, Lise Diagne, Adrien Couetoux},
pdfkeywords={Machine Learning \and Embeddings \and Real World Evidence},
}

\begin{document}
\maketitle

\begin{abstract}
In pharmaceutical R\&D, predicting the efficacy of a pharmaceutical in treating a particular disease prior to clinical testing or any real-world use has been challenging.
In this paper, we propose a flexible and modular machine learning-based approach for predicting the efficacy of an untested pharmaceutical for treating a disease.
We train a machine learning model using sets of pharmaceutical-pathway weight impact scores and patient data, which can include patient characteristics and observed clinical outcomes.
The resulting model then analyses weighted impact scores of an untested pharmaceutical across human biological molecule-protein pathways to generate a predicted efficacy value.
We demonstrate how the method works on a real-world dataset with patient treatments and outcomes, with two different weight impact score algorithms
We include methods for evaluating the generalisation performance on unseen treatments, and to characterise conditions under which the approach can be expected to be most predictive.
We discuss specific ways in which our approach can be iterated on, making it an initial framework to support future work on predicting the effect of untested drugs, leveraging RWD clinical data and drug embeddings.
\end{abstract}


\section{Introduction}

\subsection{Context}
Predicting the treatment effect of these novel molecules in advance of any clinical trial testing would allow pharmaceutical companies to focus their resources on the molecules, target populations, and trial designs with the highest likelihood of success. This could greatly reduce the time, cost, and risk to patients of drug development while increasing the chances of successfully bringing a drug to market.

Estimating the effect of treatments that have already gone through clinical trials or are already prescribed in the real world (whether for the indication and target population of interest or not) using machine learning and real-world data is already well-established (e.g., \cite{liu_real-world_2022}). However, this approach is currently limited to predicting the treatment effects of already-approved pharmaceuticals. It cannot determine the effect of novel treatments that have yet to be tested in clinical trials or real-world populations. Pre-clinical data, which is the only source of experimental data available for a novel pharmaceutical prior to clinical trials, can be helpful in surfacing signs of efficacy on animal models or cell assays, but can struggle to integrate the complexity of clinical patient characteristics and will typically focus on surrogate biomarkers rather than the actual clinical endpoint of interest. In this paper, we demonstrate methods for achieving this and discuss the relative advantages and disadvantages of possible approaches.

In order to be able to predict the effect of an untested treatment, it is necessary to be able to identify functionally relevant information about the novel drug, compare this information to that of existing drugs on the market and thus, based on their similarities and differences, infer its potential efficacy. A similar challenge has been faced in other domains where large “dictionaries” of terms need to be utilised in predictive modelling, with word embeddings being the most notable example. In these situations, it has been shown effective to create vector embeddings that are more conducive to mathematical manipulation, including integration into machine learning pipelines. Specifically, in this work, embeddings are created that: i) contain information that is possible to know about an untested treatment (e.g., molecular structure, pathways impacted), ii) contain information that it is reasonable to expect may actually be predictive of treatment effect, and iii) can be directly compared to comparable embeddings of existing treatments either in the real world or clinical trials. Under these assumptions, one can attempt to extrapolate the effect of any novel treatment based on the similarity of an embedding of its properties to the other existing treatments. Here, we demonstrate that it is possible to use molecular databases to identify information about novel treatments (e.g., interactions with proteins, chemical structure, role in molecular pathways) and generate embeddings for these treatments, which can be used as features in a predictive model to determine treatment outcomes.

Specifically, we demonstrate one approach to enriching real-world data sets using information from knowledge graphs based on information extracted from published literature or public data, although it is possible to extend the approach to also be used on proprietary experimental data that may inform the impact of a molecule on different biological pathways. A knowledge graph determines the strength of association between a target protein of a known, tested pharmaceutical and each human biological pathway. We develop an embedding which consists of a vector comprising weightings for the individual impacts of a drug on each given pathway. This is then used as a feature in a predictive model along with de-identified patient-level real-world data from patients with a disease of interest to predict what impact different pathway-weighted impacts of drugs have on patient outcomes. We can then use this model to counterfactually determine what the effect of an untested pharmaceutical is, based on its respective impact on each biological pathway.

\subsection{Related work}

Real-world data and real-world evidence have been increasingly used to quantify the effectiveness of novel treatments across various therapeutic areas \cite{wang_real-world_2021}. \citealp{ching_opportunities_2018} have applied deep learning to molecular target prediction to predict the impact of novel treatments using molecular pathways and real-world data. Furthermore, real-world studies are now well-established as sound evidence to support regulatory decisions and guide decision-making \cite{geldof_comparative_2019}. \citealp{liu_real-world_2022} highlighted real-world data's added value compared to traditional randomized clinical trial data when predicting treatment outcomes.

Molecular profiles are used extensively jointly with statistical methods to predict treatment outcomes, thereby demonstrating the relevance of molecular pathways in treatment effect estimation \cite{wu_analyses_2023}. \citealp{nakagawa_2021} have used bioinformatics databases to explore novel treatment mechanisms and identify new indications for existing molecules. 

The joint use of real-world data and bioinformatics databases has emerged in recent years. The advances in machine learning for drug response prediction have emulated the exploration of new molecule representations, including knowledge graph embeddings \cite{xin_2021}. \citealp{celebi_evaluation_2019} propose an approach to predict drug-drug interactions using knowledge graph embeddings. Their evaluation approach assesses DDI predictions using cross-validation. \cite{wen_multimodal_2023} leverage Electronic Health Records (EHR) and clinical semantics to embed patient and molecule-level data to predict drug-disease interaction.

While previous work primarily focused on real-world evidence and molecular representations separately, this work proposes an approach to treatment effect estimation encapsulating both aspects, thereby accelerating the drug development process by providing early efficacy estimates.

\subsection{Purpose}

The purpose of this paper is to introduce a novel, predictive framework for estimating the effects of untested pharmaceutical treatments prior to clinical trials or real-world application.
Our method uses the predictive power of functionally relevant information about novel treatments, notably molecular structure and impacted pathways, and compares this to existing, tested treatments.
By creating embeddings that encapsulate this information and using real-world datasets enriched with knowledge graphs, we illustrate the potential for this approach to predict patient outcomes based on varying pathway impacts of drugs.
This paper also presents a mechanism for data scientists to anticipate which treatments this predictive approach would work best for, further enabling targeted and efficient drug development strategies.

Despite its inherent limitations which we discuss later in this paper, we believe that this compact, innovative method can, subject to clinical, regulatory and ethical considerations that pertain to all in silico methods, revolutionize drug development by reducing time and cost burdens, mitigating risks to patients, and enhancing the likelihood of successful treatments reaching the market.
In particular, we designed our approach to be flexible and modular, to support future innovation and make it simple to adjust for specific indications, drugs and endpoints.
For example, one can easily implement it with different molecule embedding approaches, or different treatment effect estimation models.

\section{Materials and Methods}

\subsection{Treatment effect modelling}

Although treatment effects are eventually estimated using RCTs\footnote{Randomized controlled trials.}, real-world data provides significant complementarity (e.g.,. much larger size, often representative of the entire population without exclusion criteria), but also additional challenges (e.g.,. lack of randomisation, missing data, confounders).

Although numerous methods for treatment effect estimation exist to mitigate these challenges \cite{causal_book}, we focus on methods that allow for any arbitrary vector representation of the treatment to be plugged in (e.g.,. g-estimation \cite{wang_g-computation_2017}), i.e. methods whose success primarily depends on the quality of an underlying model that predicts treatment outcomes (known as \emph{outcome model}).

These methods estimate $\mathbb{E}\left[Y | A=a, X=x\right]$, the expected patient outcome following a new treatment, with a model $f$, conditioned on the treatment received A and the patient covariates $X$. In practice, $Y$ is the best proxy for a patient endpoint available in RWD\footnote{Real-world Data} (e.g.,. a lab measurement made 12 weeks after the start of treatment, which correlates with the actual clinical endpoint that would typically only be measured consistently in an RCT). When $A \in {0,1}$, this is a binary indication of whether a patient has been on treatment A or not. In the case of several treatments ${A_1, …, A_p}$, then we can consider $\vec{A} \in \{0,1\}^n$ as a one-hot encoding. However, as we will later generalise this representation to an arbitrary embedding, we will maintain a completely general framework where $A \in \mathbb{R}^n$, where n may be equal, smaller or larger than the total number of treatments. Although this includes the case where $A \in \{0, 1\}$, it is important for our purpose to handle complex treatment vectors in order to explore the generalisation of predictions to novel treatments. $X$ can contain any observable variable about the patient at index date, i.e. the date when the treatment starts. Under assumptions of exchangeability, positivity, consistency, no measurement error and no model misspecification, the above estimator $f$ can be used to compute unbiased treatment effects \cite{causal_book}.

Our focus in this paper is thus on fitting the treatment effect model $f$ and its predictive performance. All else being equal (e.g.,. observable confounders, selection bias), the quality of this will directly impact the accuracy of treatment effect estimates.

More specifically, we are interested in building a model $f$ to predict patient outcomes under a treatment $A=a_{novel}$, using data that only contains other treatments, i.e. training on treatments $a \in A_{train}$ such that $a_{novel} \notin A_{train}$. The goal is for the method to deliver a model $f$ that is an unbiased estimate of $\mathbb{E}\left[Y | A=a_{novel}, X=x\right]$.

%
%
%
%
%

\subsection{Molecule embeddings}

In order to generalise the learned effect of known molecules to novel ones, we encode each treatment (known and novel) into the same vector space.
This mapping between treatment and vector space can be done in several ways.

Although we only present two simple embedding approaches, our framework is compatible with any approach as long as it yields a real valued array for each drug.
And despite being simple to implement and run, our embeddings already present high-level desirable qualitative properties, e.g. with similar drugs being clustered together in embedding space.


\subsection{SMILES-based embeddings}

SMILES-based embeddings methods only consider the molecule’s structure (i.e. atoms), and ignore its known interactions with other proteins (whereas Knowledge Graphs (KG) focus mostly on the latter, and not on the structure itself). Most of the state-of-the-art solutions in this category rely on deep learning models trained on one or more molecule-level predictive tasks, and on extracting one or more specific layer’s activations to derive an embedding \cite{graph_dti}. They mostly vary by what type of molecule property they try to predict (e.g.,. toxicity, affinity with particular target, Drug-Target-Interaction, etc) and the choice of model architecture. The last few layers of these models often happen to be, empirically speaking, representative of molecule properties (i.e. similar molecules are close to one another in that space).

In this paper, we used the Galactica pre-trained model, because it has been trained (among other things) on SMILES data. To create embeddings for a given SMILES s, we proceed as follows:
\begin{enumerate}
	\item tokenise $s$, then pass the result through the decoder.
	\item extract the last layer’s activations of shape \small\verb+(len(s), 4096)+\normalsize, and mean-pool them to obtain a vector of shape 4096.
	\item apply PCA to reduce its dimension to 3.

\end{enumerate}

\begin{figure} 
	\includegraphics[width=\linewidth]{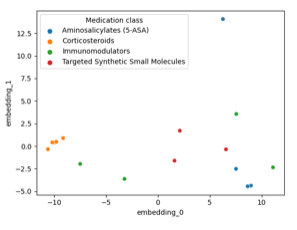}
	\caption{2-dimensional PCA for ulcerative colitis drugs using SMILES embeddings}
	\label{2d_pca}
\end{figure}

For illustrative purposes, the result of this process is shown in Figure
\ref{2d_pca} with a 2-dimensional PCA, and grouped by drug type, for all
ulcerative colitis drugs that have a SMILES.

\subsection{KEGG-based embeddings}

According to its starting page “KEGG DRUG is a comprehensive drug information resource for approved drugs in Japan, USA and Europe, unified based on the chemical structure and/or the chemical component of active ingredients. Each KEGG DRUG entry is identified by the D number and associated with KEGG original annotations including therapeutic targets, drug metabolism, and other molecular interaction network information.”

In other words KEGG is a knowledge graph that links drugs to each other via their role in diseases and a number of functional elements related to the drug or disease. While KEGG contains a fair amount of free text, the relevant data is either codified or at least standardised i.e. KEGG offers machine-accessible semantic information on drugs. This semantic information provides insight into the many different functional similarities between drugs and therefore should encode the information needed to extrapolate efficacy from tested drugs to known, but so far untested, drugs.

\subsubsection{KEGG embeddings algorithm}
While in principle, one could download KEGG into a graph database such as Neo4J and apply graph2vec or any other KG embedding method, we opted for a more lightweight approach to start:

\begin{itemize}
	\item There are numerous methods for processing a knowledge graph into embeddings, but we only aim to demonstrate the potential of our approach with a simple embedding strategy. Exploring the value of more advanced embedding strategies is left for future work.
	\item Our approach only requires embedding a tiny subset of all the drugs included in Kegg, and we chose to process only the drugs we needed to accelerate the pace of experimentation.
	\item This simple approach allows any subsequent code run to use the latest data from Kegg via their API.
\end{itemize}

Our method to compute drug embeddings from Kegg follows the steps below:
\begin{enumerate}
	\item Assemble a list of relevant drugs to analyse (in our case, linked to ulcerative colitis augmented by a few other relevant ones).
	\item For each drug in the list, extract:
	\begin{itemize}
		\item The linked diseases plus all drugs linked to these diseases
		\item The information on pathway/target codes
		\item The efficacy data
		\item The drug class/group

	\end{itemize}

	\item Concatenate all these elements into a single list per drug entry.


	\item One-hot encode the resulting codes or standardised text snippets, i.e. for each drug in the list of relevant drugs, encode their elements into counts. Each element corresponds to the number of links to the disease or pathway.
		In most cases these counts are 0 or 1, i.e. present or absent but linked drugs could appear several times if they are linked via multiple diseases.
	\end{enumerate}

Note that here we use all the drugs linked to the indication of interest that can be found in Kegg, and not only approved drugs that would be found in real-world data. This allows us to extrapolate to non-approved drugs. 

Doing so yields a relatively sparse bag-of-word-like vector (dimensionality of order $10^2$ - $10^3$), which we can further encode via TFIDF - where all drugs of interest are equivalent to the whole document.
These sparse embedding vectors can be used to calculate mutual distances between drugs.
While these could be used directly for causal modelling due to the relatively small data sets, we also apply dimensional reduction via PCA, similar to the method used for SMILES data.

\begin{figure}
	\includegraphics[width=\linewidth]{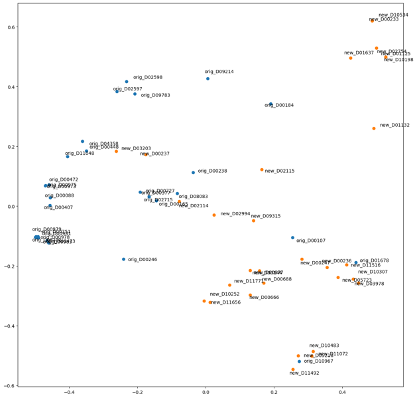}
	\caption{2-dimensional PCA for ulcerative colitis drugs using Kegg embeddings}
	\label{kegg_2d_pca}
\end{figure}

Reducing to 2 principal dimensions, we get Figure \ref{kegg_2d_pca},
where the blue drugs are those in the original list, and the orange ones are a subset of drugs linked via diseases, the latter filter not being a prerequisite, but rather only used for visibility reasons.

\subsection{Dimensionality reduction for embeddings}

A very large number of features can cause the treatment effect modelling pipeline to underperform due to overfitting.
Although feature selection is a well established research area \cite{outcome_adaptive_lasso}, we have experimented with methods designed particularly to deal with the problem at hand, i.e. select the best set of treatment-embedding features to generalise to a particular set of unseen treatments (e.g.,. ensure overlap between seen and unseen treatments in feature space).
Hence, in this paper, we focus on using Principal Component Analysis (PCA) to reduce the dimension of treatment embeddings down to a number that is adjusted following a sensitivity analysis.

\subsection{Method validation}

We ran two sets of experiments on one dataset to validate our approach and derive insights that could be applied to future deployments of this approach. This validation had the purpose of addressing two key questions:
	\begin{enumerate}
	\item How well can treatment effect be estimated, on a treatment that is unavailable to the model during training, for different types of embeddings (one-hot, SMILES-based, KEGG-based)?

	\item Are there any features of treatment embeddings that can help predict the success (or failure) of this approach?
	\end{enumerate}

\subsubsection{Comparative evaluation of different treatment embeddings in unseen-treatment outcome estimation}

We designed a process to estimate the ability of different modeling approaches to generalise to an unseen treatment.

Given a dataset X with patients who received treatments from a list \small\verb+[tx_1, ... tx_n]+\normalsize, and given a list of available treatment embeddings \small\verb+[emb_1, ..., emb_m]+\normalsize, we run the pseudo-code described in Algorithm \ref{eval_algo}.

%

\begin{figure*}[!htbp]
\begin{lstlisting}[language=Python, caption = Logic for comparative evaluation of treatment embeddings, label={eval_algo}]
for i in range(N_bootstrap):
  # Bootstrap the entire data with replacement
  X_boot = resample(X, with_replacement=True)

  for unseen_treatment in [tx_1, ..., tx_n]:
    # For each treatment, train on all other treatments, and evaluate on that treatment
    train_data = X_boot[X_boot["treatment"]!=unseen_treatment]
    test_data = X_boot[X_boot["treatment"]==unseen_treatment]

    baseline_model = train_without_treatment(train_data)

    for treat_embedding in [emb_1, ..., emb_m]:
      # Build and evaluate a model for each embedding to test
      treatment_model = train_model(train_data, treatment_embedding)
      performance_metrics[i, unseen_treatment, treat_embedding] = evaluate(treatment_model, baseline_model, test_data)
\end{lstlisting}
\end{figure*}

%
%
%

This results in a set of performance metrics for each combination of bootstrap iteration, unseen treatment and treatment embedding,

The performance metrics can be adjusted to be relevant to the task at hand, e.g.,. regression or classification metrics. For example, if treatment outcomes are values between 0 and 100, modelled as a regression, one can use Mean Squared Error (MSE) and use the above routine to compute the error per bootstrap iteration, unseen treatment and embedding type.

To enable the comparison of multiple treatment embedding approaches against a single scale, we compare each of them to the same baseline model. We use as a baseline model the one that does not use any treatment variable, and only uses patient covariates (e.g.,. demographics, past treatments, labs).

Assuming a metric where lower is better like MSE, the boolean flag that indicates whether or not a specific embedding approach outperformed the baseline model, for a given bootstrap iteration and unseen treatment is equal to:
\begin{itemize}
	\item 1 if the performance metric of the model using the selected embedding type is higher than the baseline model.
	\item 0 otherwise.
\end{itemize}

\subsubsection{Meta-analysis to link treatment embedding to predicted success of the approach}
Once the above analysis is complete, one can use its results to test hypotheses that could link properties of a treatment (e.g.,. how “novel” it is compared to known treatments) to the success of the modeling approach (e.g.,. how often a predictive model using treatment embeddings outperforms a model that uses none).

We use the boolean flag that indicates whether or not a an embedding approach outperformed the baseline model
as the dependent variable, and test a selected number of embedding-related features as covariates. These features represent the information that would be available to practitioners based on treatment embeddings only, and represent the position of a specific unseen treatment in embedding space compared to all other treatments available for training.

An example of such a covariate is $$\min\limits_{t_{x} \in train\_treatments} \|t_{x} -\mathrm{unseen\_treatment}\|$$ i.e. the euclidean distance between the unseen treatment and its nearest neighbour out of the treatments used for training (all measured in embedding space).

Depending on the kind of model used, different levels of insights can be derived.

\section{Results and discussion}
\subsection{Data}

We base our analysis on:
\begin{enumerate}
	\item A dataset containing comprehensive collection of US claims data sourced from various providers, encompassing both medical (primary and secondary care visits, procedures) and pharmacy claims (Komodo Health). The dataset captured patient demographic data (date of birth, gender, location), and medical events from 2015 onwards (including diagnoses (primarily ICD10), procedures (CPT, HCPCS, or ICD-10-PCS), prescriptions (NDC11 codes, days of supply, quantity dispensed), HCP specialty, insurance coverage plans, claims cost, visit types - including hospital and physician visits). This data is used to build and evaluate treatment effect models.
	\item A list of treatments and basic information. This contains the list of treatments to analyse, along with treatment information that is used to compute embeddings (e.g.,. SMILES and treatment name to map them to external data sources).
	\item Kegg Knowledge Graph. This contains more detailed information about treatments, their relationship to proteins, targets, indications, pathways and other drugs.
\end{enumerate}

\subsection{Models}

We use the \small\verb+xgboost.XGBRFRegressor+ \normalsize estimator to train all the models, using \small\verb+n_estimators=300+\normalsize, \small\verb+colsample_bynode=0.6+ \normalsize and \small\verb+min_child_weight=0.001+ \normalsize as training parameters. Each model goes through a feature selection procedure using the \small\verb+SelectFromModel+ \normalsize scikit-learn selector to select 20 features.

%

\subsection{RWD features and proxy endpoint}
\subsubsection{Treatment blocks}
In order to train the model, we create a master table with the unit of analysis being a patient-treatment block. A treatment block can be inferred from datasets containing information about doctor visits, pharmacy collection data, and other information about treatments (e.g., drug washout periods). By chaining these events together, one can infer the start date and end date of the treatment, thus creating a treatment block. Note that a patient may have multiple treatment blocks in their full treatment journey, thus the same patient can appear multiple times in the master table with different treatment blocks.

\subsubsection{Features}
Once the patient-treatment blocks are defined, we generate features at time of prediction that are related to  i) patient characteristics (e.g., age, gender), ii) medical diagnoses (e.g., presence of certain indications) and iii) treatments (e.g., use of certain steroids). For (ii) and (iii) we also generate lookback features, which can be both binary or a count. For example, one feature could describe “presence of diagnosis X at time of prediction”, and another could be “presence of diagnosis X in the last 6 months”. For some features a lifetime window is taken, such that the feature becomes “presence of diagnosis X in this patient’s history”.

\subsubsection{Target variable}
At this point, the goal is to come up with a prediction target (endpoint) that could be indicative of a treatment being effective or not. Based on the available data we built the target as the percentage of treatment block duration that a patient has spent on steroids.
This can be seen as an appropriate endpoint from a medical perspective, as we would expect (under proper controls), that as effectiveness of a treatment increases, the required steroid administration should decrease as a consequence of reduced flare-ups in the patient journey. It is important to control for events other than the treatment efficacy that could lead to steroid administration, hence we include short and long-term presence of inflammatory diseases that may be treated with steroids (e.g., bronchitis and panniculitis) as features in our master table.

\begin{figure*}[!htbp]
	\includegraphics[width=\linewidth]{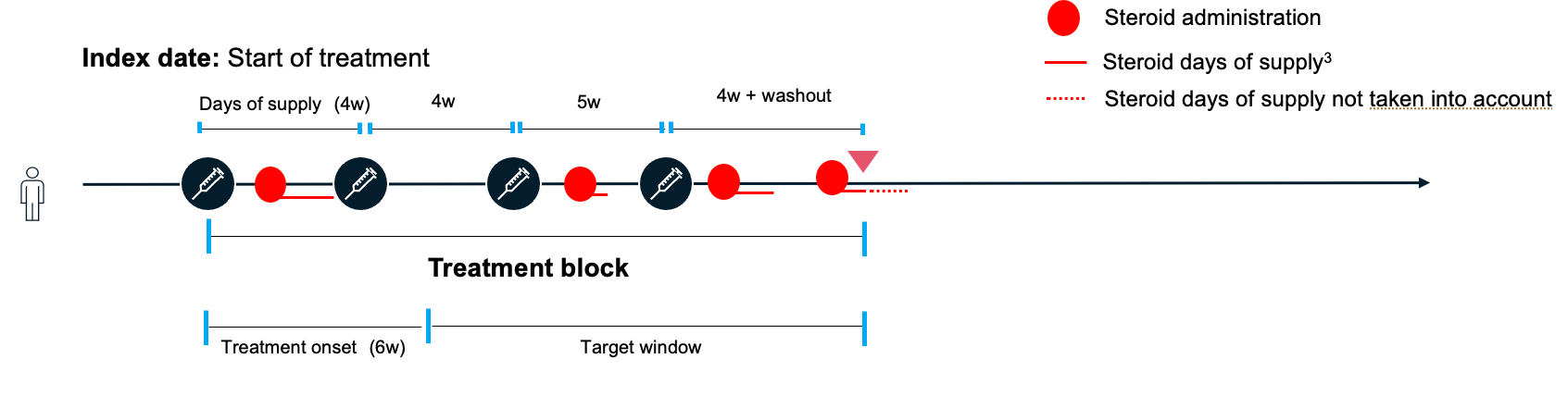}
	\caption{Description of the unit of analysis.}
	\label{uoa}
\end{figure*}

In the example shown in Figure \ref{uoa}, note how there are a number of steroid administrations present within the treatment block, each with their own duration (inferred through the days of supply found in the data). Note how the first steroid administration falls within the “treatment onset window”. It is important that this steroid administration is not counted towards the target, as we want a sufficient amount of time to pass after commencing the treatment before inferring the efficacy through steroid usage.

We can now calculate the target as:

Target = Total steroid days of supply in target window / Total duration of steroid block.

After repeating the above process for all treatment blocks, the target distribution looks as shown in appendix \ref{appendix:target_dist}.

\subsubsection{Alternative target variables}
There are a number of alternative target variables that one could choose as an appropriate endpoint. While some may be medically appropriate, not all can be found in any given real-world dataset (e.g., lab test results for the inflammatory biomarker C-reactive protein).

Other possible target variables that one could use, and which can be inferred from RWD, is a binary steroid flag (although one needs to appropriately control for treatment duration so as not to bias the predictions towards long treatment windows), change in treatment to an alternate therapy, procedures that suggest worsening of condition (e.g.,. colectomy/colostomy/proctocolectomy, presence of diagnoses related to complications of UC, presence of immunosuppression and presence of hospital admission).

For this work, the percentage of a treatment block spent on steroids was chosen as it i) was available in the RWD, ii) appropriately controls for treatment block duration, and iii) had good coverage across the cohort (and thus good coverage across treatmentsx).

\subsection{Ulcerative Colitis treatments}
In addition to real-world data which contains observations of patient events (including treatments), we use a treatment table to store information specific to each treatment included in our analysis. This additional table is used to link each treatment to external data or other methodology that is able to produce treatment embeddings (e.g.,. models pre-trained on SMILES, Knowledge Graph with a specific drug ID).

This table is available in appendix \ref{appendix:uc_treatments}.

\subsection{Comparative performance of different embeddings to novel treatment outcome prediction}
We implemented the evaluation routine described in the previous section, with:
\begin{itemize}
	\item 14 treatments (see appendix \ref{appendix:uc_treatments} for the full list); each one used as an “unseen treatment” once per bootstrap iteration
	\item 10 bootstrap iterations
	\item 4 models in total - baseline with no treatment variable at all, one-hot, smiles-based and Kegg-based.
	\item MSE as the key model performance metric, i.e. 40\% means that 40\% of bootstrap iterations lead to a model having a lower MSE than the baseline model on that same bootstrap iteration.
\end{itemize}

\begin{table*} 
	\caption{\% of bootstraps where a given embedding method yielded better model performance than the baseline}
	\centering 
	\begin{tabular}{lllll}
	\toprule
	{} & Unseen treatment & One-hot encoding & SMILES-based embeddings & Kegg-based embeddings \\
	\midrule
	0  &       adalimumab &            0.00\% &                   0.00\% &                30.00\% \\
	1  &     azathioprine &           40.00\% &                   0.00\% &                20.00\% \\
	2  &      balsalazide &           10.00\% &                  90.00\% &                 0.00\% \\
	3  &     cyclosporine &            0.00\% &                  20.00\% &                40.00\% \\
	4  &        golimumab &           10.00\% &                   0.00\% &                80.00\% \\
	5  &       infliximab &          100.00\% &                   0.00\% &                30.00\% \\
	6  &   mercaptopurine &           70.00\% &                   0.00\% &                40.00\% \\
	7  &       mesalazine &           10.00\% &                  20.00\% &                20.00\% \\
	8  &       olsalazine &           60.00\% &                  20.00\% &                30.00\% \\
	9  &    sulfasalazine &           50.00\% &                 100.00\% &                 0.00\% \\
	10 &       tacrolimus &           30.00\% &                  70.00\% &               100.00\% \\
	11 &      tofacitinib &           50.00\% &                  70.00\% &                90.00\% \\
	12 &      ustekinumab &            0.00\% &                   0.00\% &                 0.00\% \\
	13 &      vedolizumab &            0.00\% &                   0.00\% &                60.00\% \\
	\bottomrule
	\end{tabular}
	\label{bootstrap}
\end{table*}

Our results shown in Table \ref{bootstrap} indicate strong variations between both treatments and embedding methods, i.e. the best performing method varies across treatments, and for a given method, absolute performance varies by treatment.

In our experiment, we observe two seemingly counterintuitive results.
First, for some treatments, the baseline model outperforms all others, and second, the one-hot encoded model sometimes outperforms all others.
In the first case, this may indicate test treatments that none of the embeddings accurately capture.
In other words, the treatment variable may be helpful, but the test treatment could be too different from train treatments to be generalised to.
In the second case, this may indicate test treatments that happen to match the 0-encoded treatment, i.e. the train treatment encoded entirely as zeros.
When this happens, the test treatment is seen by the model exactly as the 0-encoded treatment, which may lead to high performance when these two treatments have similar effect.


\subsection{Meta-analysis to link embeddings to likelihood of success in predicting novel treatment outcomes}
The previous analysis shows that there appears to be a link between which treatment is kept hidden and the performance of models with treatment embeddings compared to the baseline model.

In this analysis, we measure the validity of this hypothesis. We investigate the link between distance metrics on an unseen treatment (compared to other treatments) and the performance of a model when applied to that unseen treatment.

Our unit of analysis is a \small$(\mathrm{bootstrap\_iteration, unseen\_treatment})$ \normalsize tuple, with the dependent variable being a binary flag to indicate if the Kegg-based model outperformed the baseline on that given tuple.

To measure the novelty of an unseen treatment, we considered two metrics, and two distance measurements (euclidean and cosine), resulting in four different treatment-level features in total. The two metrics we tested were:
\begin{itemize}
	\item Distance between a novel treatment \small$\mathrm{t_{x}\_novel}$ \normalsize and its nearest neighbour in known treatments, i.e.\small$\min\limits_{t_{x} \in \mathrm{train\_treatments}} \mathrm{(distance(t_{x}, unseen\_treatment))}$\normalsize.
	\item Distance between a novel treatment \small$\mathrm{t_{x}\_novel}$ \normalsize and \small$mean(tx \in \mathrm{all\_treatments})$ \normalsize i.e. the centre of gravity of all treatments available (including the novel treatment).

\end{itemize}

This results in an input table with one dependent variable \small$\mathrm{ft\_kegg\_embeddings\_perf\_higher\_than\_ft\_no\_treatment}$ \normalsize and four different possible covariates. A sample of this table is available in Appendix \ref{appendix:sample}.

\begin{figure*}[h]
	\includegraphics[width=\linewidth]{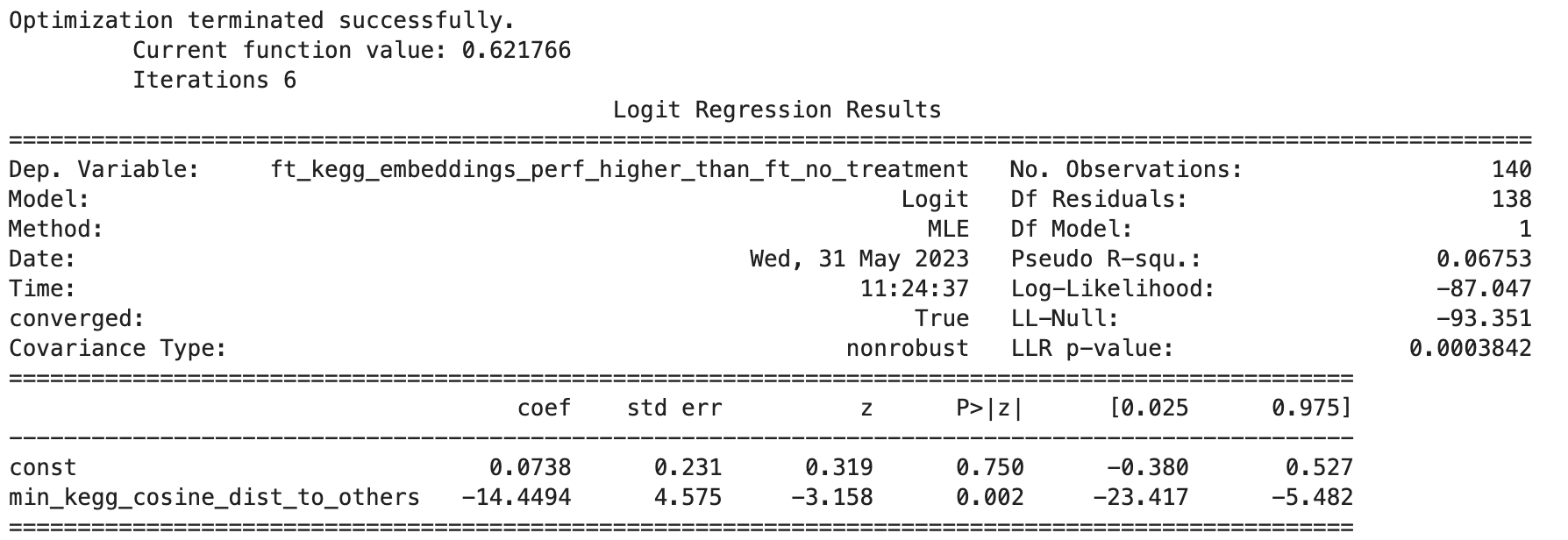}
	\caption{Regression of performance on the minimum of cosine distance to known treatments. The target variable is a binary flag indicating if the Kegg-based model outperformed the baseline on that given tuple.}
	\label{min_cos_dist}
\end{figure*}

The results of the logistic regression for the minimum cosine distance to the novel treatment are shown in Figure \ref{min_cos_dist}. The other three regressions performed on the minimum euclidean distance, the cosine distance to the mean treatment and the euclidean distance to the average treatment are also available in Appendices \ref{appendix:min_euclidean_dist}, \ref{appendix:cos_dist_to_mean} and \ref{appendix:euclidean_dist_to_mean}.

Our results show that, for each distance metric, there is a significant relationship between that distance and the Kegg-based model outperforming baseline, and that relationship is always a negative coefficient.
In other words, the larger the distance between a novel treatment and known treatments is, the lower the chances of outperforming baseline are.
Conversely, the closer a novel treatment is to a known treatment in embedding space, the more likely it is for our method to outperform baseline.

\section{Limitations}
Our proposed approach is subject to all existing limitations of treatment outcome prediction on observational data, e.g.,. non-randomised and biased treatment allocation, unobserved confounders, or informative missingness.
The novelty of our approach is in how it estimates the effect of an unobserved treatment based on observations of other treatments.

One of the main limitations is that better performance than a treatment-less baseline can only be obtained if the novel treatment overlaps to some extent with known treatments in the embedding space used to encode treatments.
Our analysis indicates a negative relationship between i) the distance between novel treatment and known treatments, and ii) predictive power of the resulting model on the novel treatment.
If one cannot embed a novel treatment in a space where that treatment is close to known treatments, our method is unlikely to yield better results than simply ignoring the treatment allocation entirely.

Critically, the treatment embedding approaches define what type of information about treatments is captured.
The methods we proposed capture molecule structure in one case (based on SMILES) and the relationship of the treatment with indications, pathways and other drugs in the other.
There are some clinically relevant treatment features that none of these methods capture, such as a drug being topical or systemic.
The possible approaches to encode treatments as vectors are virtually infinite, and significant progress could be achieved by improving on the embedding methods we proposed.

\section{Future work}
Beyond its results on one real-world dataset, our approach sets up a framework with significant potential and clear paths for improvement.
Our approach can be adjusted to work with (i) different drug embedding approaches, (ii) different treatment effect estimation, and (iii) different evaluation metrics (e.g. with a focus on estimating treatment effect accurately rather than patient outcomes).

To embed drugs, listing all possible approaches is beyond the score of this paper, but we expect knowledge graph based methods, like node2vec\cite{grohe2020word2vec}, to be particularly effective, as this is where most of the data can be found for untested drugs, and as it contains information about pathways, indications and targets.
Other approaches, e.g. deep neural network based, can equally be used within this paper's framework.

On treatment effect estimation method, this paper focused on simply predicting patient outcomes, with a traditional machine learning model.
For this application, it is more common to go a step further to also estimate the difference between a novel treatment effect and other existing treatments.
This can be done with methods such as g-estimation \cite{wang_g-computation_2017} or Double Machine Learning\cite{Chernozhukov2018}.

Lastly, we tested the method on one dataset, looking only at MSE as a proxy for the quality of treatment effect estimation.
However, one could evaluate the approach not only on other datasets (e.g., other endpoints, indications) but also on different metrics better suited to evaluate causal effect estimators like R-risk \cite{Doutreligne_giaf016}.
Additionally, one could create synthetic data, with known hidden treatment effects, to better understand how and when the method works best.


%

\section{Conclusion}
We have proposed a method for estimating the effect of novel treatments, based on observed patient outcomes for existing treatments.
Our method relies on embedding treatments into a vector space where closeness reflects similarity of treatments.

We proposed two different methods to embed treatments, one based on SMILES to capture molecule structure, and another based on Kegg’s knowledge graph that captures relationships between drugs, proteins, target, pathways and indications.

We evaluated our approach on a real-world dataset, and reported its performance compared to a benchmark with no treatment information, for both embedding approaches.

Finally, we identified a negative relationship between the distance from a novel treatment to known treatments in embedding space, and the success of our method.
This indicates a set of metrics that practitioners could use to define the right embedding methodology for their own dataset, i.e. a methodology such that it minimises the distance between the novel treatment and treatments available in patient data.


Our proposed approach unlocks the potential to provide accurate estimates of the effect of a drug early in the drug development chain and support internal and regulatory decision-making by accelerating access for patients in urgent need of a novel treatment.

\bibliographystyle{unsrtnat}
\bibliography{references}  






\newpage
\appendix
\onecolumn
\section{Distribution of the target variable}
\label{appendix:target_dist}
The target is defined as Total steroid days of supply in target window / Total duration of steroid block.
The total steroid days of supply in target window does not include the treatment onset period.
\begin{figure}[!htbp]
	\centering
	\includegraphics[scale=0.5]{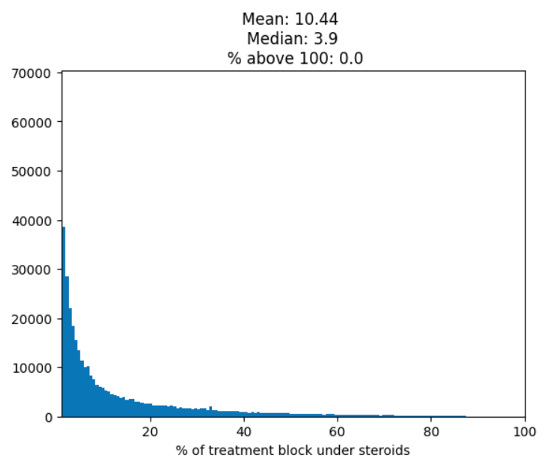}
	\caption{Distribution of the target variable.}
\end{figure}

\newpage
\section{Table of ulcerative colitis treatments present in the real-world dataset}
\label{appendix:uc_treatments}
\begin{table*}[h] 
\resizebox{\textwidth}{!}
{\begin{tabular}{lllll}
\toprule
{} &        Generic name &                    Medication class & KEGG\_code &                                                                                                                                                                                                                        SMILES \\
\midrule
0  &       Sulfasalazine &            Aminosalicylates (5-ASA) &    D00448 &                                                                                                                                                                   OC(=O)C1=CC(=CC=C1O)\textbackslash N=N\textbackslash C1=CC=C(C=C1)S(=O)(=O)NC1=NC=CC=C1 \\
1  &          Mesalamine &            Aminosalicylates (5-ASA) &    D00377 &                                                                                                                                                                                                       NC1=CC(C(O)=O)=C(O)C=C1 \\
2  &          Olsalazine &            Aminosalicylates (5-ASA) &    D00727 &                                                                                                                                                                               OC(=O)C1=CC(=CC=C1O)\textbackslash N=N\textbackslash C1=CC=C(O)C(=C1)C(O)=O \\
3  &         Balsalazide &            Aminosalicylates (5-ASA) &    D02715 &                                                                                                                                                                        OC(=O)CCNC(=O)C1=CC=C(C=C1)\textbackslash N=N\textbackslash C1=CC=C(O)C(=C1)C(O)=O \\
4  &          Prednisone &                     Corticosteroids &    D00473 &                                                                                                                                        [H][C@@]12CC[C@](O)(C(=O)CO)[C@@]1(C)CC(=O)[C@@]1([H])[C@@]2([H])CCC2=CC(=O)C=C[C@]12C \\
5  &        Prednisolone &                     Corticosteroids &    D00472 &                                                                                                                                     [H][C@@]12CC[C@](O)(C(=O)CO)[C@@]1(C)C[C@H](O)[C@@]1([H])[C@@]2([H])CCC2=CC(=O)C=C[C@]12C \\
6  &  Methylprednisolone &                     Corticosteroids &    D00407 &                                                                                                                              [H][C@@]12CC[C@](O)(C(=O)CO)[C@@]1(C)C[C@H](O)[C@@]1([H])[C@@]2([H])C[C@H](C)C2=CC(=O)C=C[C@]12C \\
7  &          Budesonide &                     Corticosteroids &    D00246 &                                                                                                                         [H][C@@]12C[C@H]3OC(CCC)O[C@@]3(C(=O)CO)[C@@]1(C)C[C@H](O)[C@@]1([H])[C@@]2([H])CCC2=CC(=O)C=C[C@]12C \\
8  &      Hydrocortisone &                     Corticosteroids &    D00088 &                                                                                                                                      [H][C@@]12CC[C@](O)(C(=O)CO)[C@@]1(C)C[C@H](O)[C@@]1([H])[C@@]2([H])CCC2=CC(=O)CC[C@]12C \\
9  &        Azathioprine &                    Immunomodulators &    D00238 &                                                                                                                                                                                    CN1C=NC(=C1SC1=NC=NC2=C1NC=N2)[N+]([O-])=O \\
10 &    6-mercaptopurine &                    Immunomodulators &    D01678 &                                                                                                                                                                                                            S=C1N=CNC2=C1NC=N2 \\
11 &        Cyclosporine &                    Immunomodulators &    D00184 &  CC[C@@H]1NC(=O)[C@H]([C@H](O)[C@H](C)C\textbackslash C=C\textbackslash C)N(C)C(=O)[C@H](C(C)C)N(C)C(=O)[C@H](CC(C)C)N(C)C(=O)[C@H](CC(C)C)N(C)C(=O)[C@@H](C)NC(=O)[C@H](C)NC(=O)[C@H](CC(C)C)N(C)C(=O)[C@@H](NC(=O)[C@H](CC(C)C)N(C)C(=O)CN(C)C1=O)C(C)C \\
12 &          Tacrolimus &                    Immunomodulators &    D00107 &                                                         CO[C@@H]1C[C@@H](CC[C@H]1O)\textbackslash C=C(/C)[C@H]1OC(=O)[C@@H]2CCCCN2C(=O)C(=O)[C@]2(O)O[C@@H]([C@H](C[C@H]2C)OC)[C@H](C[C@@H](C)C\textbackslash C(C)=C\textbackslash [C@@H](CC=C)C(=O)C[C@H](O)[C@H]1C)OC \\
13 &            Ozanimod &  Targeted Synthetic Small Molecules &    D10967 &                                                                                                                                                                 CC(C)OC1=C(C=C(C=C1)C1=NC(=NO1)C1=C2CC[C@H](NCCO)C2=CC=C1)C\#N \\
14 &         Tofacitinib &  Targeted Synthetic Small Molecules &    D09783 &                                                                                                                                                                           C[C@@H]1CCN(C[C@@H]1N(C)C1=NC=NC2=C1C=CN2)C(=O)CC\#N \\
15 &        Upadacitinib &  Targeted Synthetic Small Molecules &    D11048 &                                                                                                                                                                 CC[C@@H]1CN(C[C@@H]1C1=CN=C2C=NC3=C(C=CN3)N12)C(=O)NCC(F)(F)F \\
16 &          Adalimumab &                           Biologics &    D02597 &                                                                                                                                                                                                                           NaN \\
17 &           Golimumab &                           Biologics &    D04358 &                                                                                                                                                                                                                           NaN \\
18 &          Infliximab &                           Biologics &    D02598 &                                                                                                                                                                                                                           NaN \\
19 &         Ustekinumab &                           Biologics &    D09214 &                                                                                                                                                                                                                           NaN \\
20 &         Vedolizumab &                           Biologics &    D08083 &                                                                                                                                                                                                                           NaN \\
21 &     Infliximab-abda &                         Biosimilars &    D02598 &                                                                                                                                                                                                                           NaN \\
22 &     Infliximab-axxq &                         Biosimilars &    D02598 &                                                                                                                                                                                                                           NaN \\
23 &     Infliximab-dyyb &                         Biosimilars &    D02598 &                                                                                                                                                                                                                           NaN \\
24 &     Infliximab-qbtx &                         Biosimilars &    D02598 &                                                                                                                                                                                                                           NaN \\
25 &     Adalimumab-atto &                         Biosimilars &       NaN &                                                                                                                                                                                                                           NaN \\
\bottomrule
\end{tabular}}
	\end{table*}

\section{Meta-analysis input table}
The link between model performance and the overlap between treatments in the embedding space is shown using the table below as input data.
\label{appendix:sample}
\begin{table*}[h] 
\resizebox{\textwidth}{!}
{\begin{tabular}{lllrrrrr}
\toprule
{} & iteration & unseen\_treatment &  ft\_kegg\_embeddings\_perf\_higher\_than\_ft\_no\_treatment &  min\_kegg\_eucl\_dist\_to\_others &  min\_kegg\_cosine\_dist\_to\_others &  kegg\_eucl\_dist\_to\_mean &  kegg\_cosine\_dist\_to\_mean \\
\midrule
0  &         0 &       adalimumab &                                                  0 &                      0.132714 &                        0.067065 &                0.449288 &                  1.644321 \\
1  &         0 &     azathioprine &                                                  0 &                      0.020659 &                        0.000384 &                0.451267 &                  0.080567 \\
2  &         0 &      balsalazide &                                                  0 &                      0.331173 &                        0.172079 &                0.580841 &                  0.760128 \\
3  &         0 &     cyclosporine &                                                  0 &                      0.072214 &                        0.006781 &                0.376017 &                  0.021591 \\
4  &         0 &        golimumab &                                                  1 &                      0.205942 &                        0.011710 &                0.201947 &                  0.034937 \\
5  &         0 &       infliximab &                                                  1 &                      0.132714 &                        0.067065 &                0.318102 &                  1.339686 \\
6  &         0 &   mercaptopurine &                                                  0 &                      0.205942 &                        0.011710 &                0.031889 &                  0.007009 \\
7  &         0 &       mesalazine &                                                  0 &                      0.078266 &                        0.007896 &                0.379357 &                  0.117350 \\
8  &         0 &       olsalazine &                                                  1 &                      0.168396 &                        0.040963 &                0.341720 &                  0.264969 \\
9  &         0 &    sulfasalazine &                                                  0 &                      0.078266 &                        0.007896 &                0.336018 &                  0.092347 \\
10 &         0 &       tacrolimus &                                                  1 &                      0.020659 &                        0.000384 &                0.463986 &                  0.090972 \\
11 &         0 &      tofacitinib &                                                  1 &                      0.072214 &                        0.006781 &                0.388419 &                  0.022621 \\
12 &         0 &      ustekinumab &                                                  0 &                      0.402923 &                        0.102162 &                0.691087 &                  1.314353 \\
13 &         0 &      vedolizumab &                                                  1 &                      0.256478 &                        0.095692 &                0.378641 &                  0.301698 \\
14 &         1 &       adalimumab &                                                  1 &                      0.132714 &                        0.067065 &                0.449288 &                  1.644321 \\
15 &         1 &     azathioprine &                                                  0 &                      0.020659 &                        0.000384 &                0.451267 &                  0.080567 \\
16 &         1 &      balsalazide &                                                  0 &                      0.331173 &                        0.172079 &                0.580841 &                  0.760128 \\
17 &         1 &     cyclosporine &                                                  1 &                      0.072214 &                        0.006781 &                0.376017 &                  0.021591 \\
18 &         1 &        golimumab &                                                  1 &                      0.205942 &                        0.011710 &                0.201947 &                  0.034937 \\
19 &         1 &       infliximab &                                                  0 &                      0.132714 &                        0.067065 &                0.318102 &                  1.339686 \\
\bottomrule
\end{tabular}}

\end{table*}

\newpage
\section{Meta-analysis logistic regression results}
\subsection{Regression of higher-than-baseline performance indicator on the minimum euclidean distance between the hidden treatment and the known treatments}
\label{appendix:min_euclidean_dist}
\begin{figure*}[hbt!]
	\includegraphics[width=\linewidth]{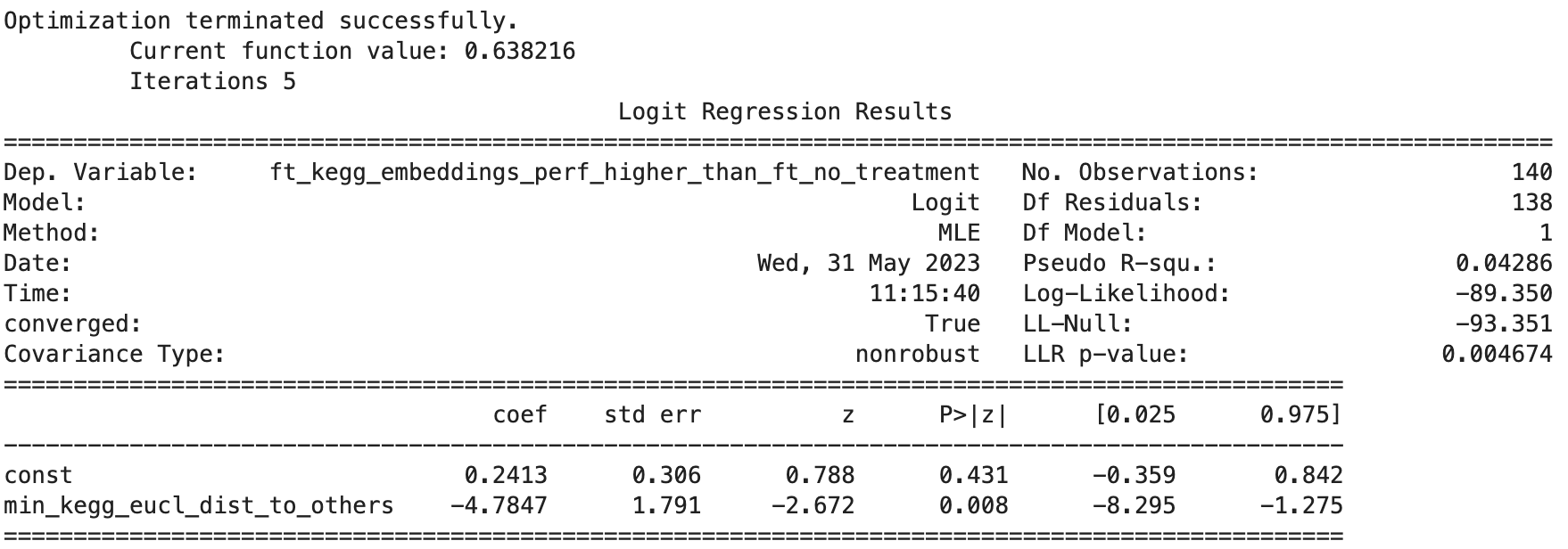}
	\caption{Minimum of euclidean distance to known treatments}
\end{figure*}

\subsection{Regression of higher-than-baseline performance indicator on the  cosine distance between the hidden treatment and the center of gravity of all known treatments}
\label{appendix:cos_dist_to_mean}
\begin{figure*}[hbt!]
	\includegraphics[width=\linewidth]{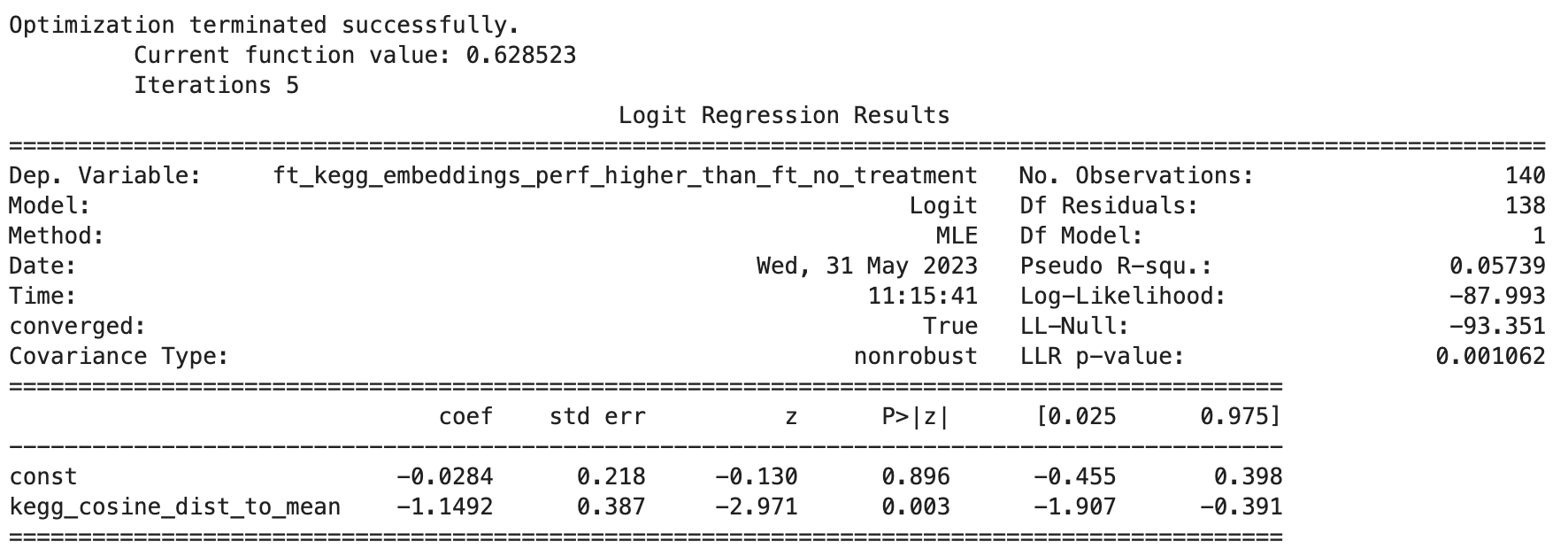}
	\caption{Cosine distance to mean of treatments}
\end{figure*}

\newpage
\subsection{Regression of higher-than-baseline performance indicator on the  euclidean distance between the hidden treatment and the center of gravity of all known treatments}
\label{appendix:euclidean_dist_to_mean}
\begin{figure*}[hbt!]
	\includegraphics[width=\linewidth]{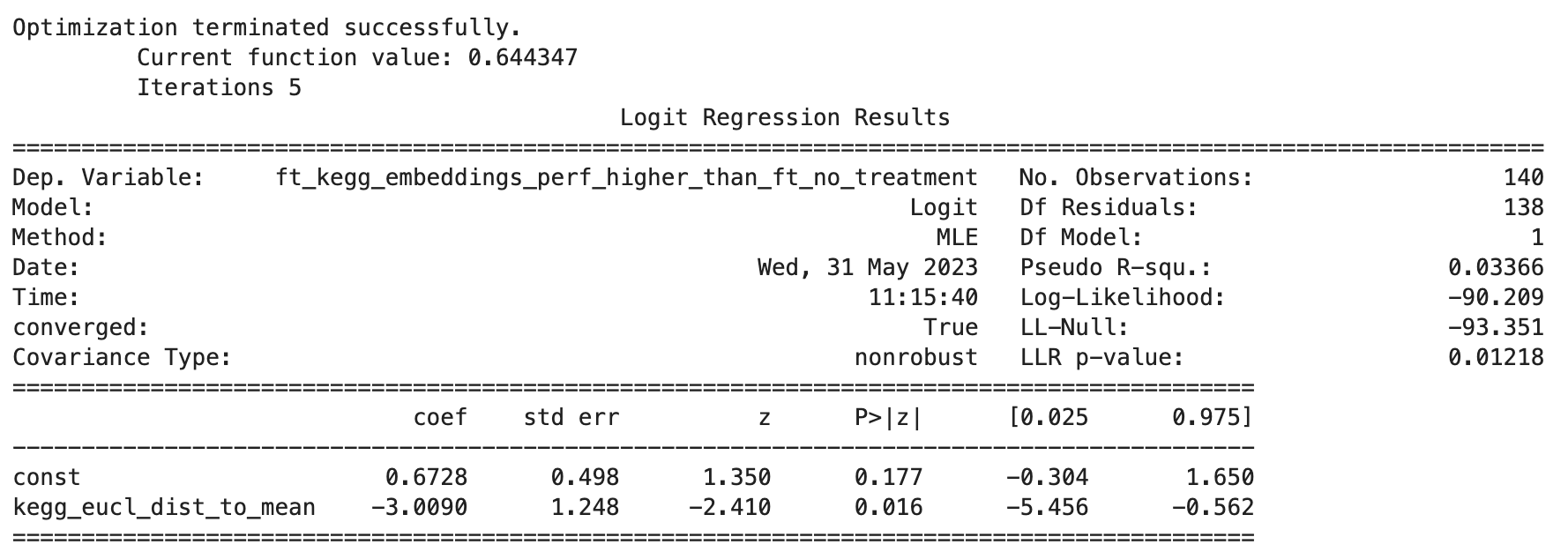}
	\caption{Euclidean distance to mean of treatments}
\end{figure*}

\end{document}